\documentclass{article}

\usepackage[final, nonatbib]{nips_2017}
\usepackage[nolist]{acronym}
\usepackage{geometry}
\usepackage{amsmath}

\usepackage{threeparttable}
\usepackage{makecell}
\usepackage{pifont}
\usepackage{hyperref}
\usepackage{graphicx}
\usepackage{caption}
\usepackage{subcaption}
\usepackage{adjustbox}
\usepackage{tabularx}
\usepackage{multirow}
\usepackage{array}
\usepackage{xcolor}
\newcolumntype{R}[2]{%
    >{\adjustbox{angle=#1,lap=\width-(#2)}\bgroup}%
    l%
    <{\egroup}%
}
\newcommand*\rot{\multicolumn{1}{R{30}{1em}}}

\newcolumntype{Z}[0]{>{\centering\arraybackslash}X}%
\newcolumntype{s}[0]{>{\hsize=.4\hsize}Z}%
\newcolumntype{n}[0]{>{\hsize=.8\hsize}Z}%

\newcommand{\cmark}{\ding{51}}%
\newcommand{\xmark}{\ding{55}}%

\begin{acronym}
\acro{RNN}{recurrent neural network}
\acro{GRU}{gated recurrent unit}
\acro{LSTM}{long short-term memory}
\acro{MFCC}{Mel-frequency cepstral coefficient}
\acro{PEM}{per-epoch noise mixing}
\acro{SNR}{signal-to-noise ratio}
\acro{ASR}{automatic speech recognition}
\acro{CNN}{convolutional neural network}
\acro{DNN}{deep neural network}
\acro{FNN}{feedforward neural network}
\acro{ACCAN}{accordion annealing}
\acro{WSJ}{Wall Street Journal}
\acro{CTC}{Connectionist Temporal Classification}
\acro{WFST}{Weighted Finite State Transducer}
\acro{WER}{word error rate}
\acro{CER}{character error rate}
\acro{LVCSR}{large-vocabulary continuous speech recognition}
\acro{ROI}{range of interest}
\acro{STAN}{sensor transformation attention network}
\acro{SER}{sequence error rate}
\acro{HMM}{hidden Markov model}
\acro{LER}{label error rate}
\acro{sMBR}{state-level minimum Bayes risk}
\acro{MLLT}{maximum likelihood linear transform}
\acro{fMLLR}{feature-space maximum likelihood linear regression}
\acro{MVDR}{minimum variance distortionless response}
\acro{STFT}{short-time Fourier transform}
\acro{ATTACC}{attention accuracy}
\acro{ATTCORR}{attention correlation}
\acro{CFE}{convolutional front-end}
\acro{BLSTM}{bidirectional long short-term memory}
\acro{E2E}{end-to-end}
\acro{DNN-HMM}{deep neural network - hidden Markov model}

\end{acronym}

\usepackage[utf8]{inputenc} 
\usepackage[T1]{fontenc}    
\usepackage{hyperref}       
\usepackage{url}            
\usepackage{booktabs}       
\usepackage{amsfonts}       
\usepackage{nicefrac}       
\usepackage{microtype}      

\title{LSTM Benchmarks for Deep Learning Frameworks}

\author{
  Stefan Braun\\
  \texttt{sb9911@gmail.com} \\
}

\begin{document}

\maketitle

\begin{abstract}
This study provides benchmarks for different implementations of \ac{LSTM} units between the deep learning frameworks PyTorch, TensorFlow, Lasagne and Keras. The comparison includes cuDNN \acp{LSTM}, fused \ac{LSTM} variants and less optimized, but more flexible \ac{LSTM} implementations. The benchmarks reflect two typical scenarios for automatic speech recognition, notably continuous speech recognition and isolated digit recognition. These scenarios cover input sequences of fixed and variable length as well as the loss functions \ac{CTC} and cross entropy. Additionally, a comparison between four different PyTorch versions is included. The code is available online \url{https://github.com/stefbraun/rnn\_benchmarks}.
\end{abstract}

\section{Introduction}
In recent years, deep learning frameworks such as PyTorch~\cite{paszke2017automatic}, TensorFlow~\cite{tensorflow2015-whitepaper}, Theano-based Lasagne~\cite{theano, lasagne}, Keras~\cite{chollet2015keras}, Chainer~\cite{chainer_learningsys2015} and others~\cite{wikipedia} have been introduced and developed at a rapid pace. These frameworks provide neural network units, cost functions and optimizers to assemble and train neural network models. In typical research applications, network units may be combined, modified or new network units may be introduced. In all cases the training time is a crucial factor during the experimental evaluation: faster training allows for more experiments, larger datasets or reduced time-to-results. Therefore, it is advantageous to identify deep learning frameworks that allow for fast experimentation. 

This study focuses on the benchmarking of the widely-used \ac{LSTM} cell~\cite{hochreiter1997long} that is available in the mentioned frameworks. The \ac{LSTM} architecture allows for various optimization steps such as increased parallelism, fusion of point-wise operations and others \cite{appleyard2016optimizing}. While an optimized \ac{LSTM} implementation trains faster, it is typically more difficult to implement (e.g. writing of custom CUDA kernels) or modify (e.g. exploring new cell variants, adding normalization etc.). Resultingly, some frameworks provide multiple \ac{LSTM} implementations that differ in training speed and flexibility towards modification. For example, TensorFlow offers 5 \ac{LSTM} variants: (1) \texttt{BasicLSTMCell}, (2) \texttt{LSTMCell}, (3) \texttt{LSTMBlockCell}, (4) \texttt{LSTMBlockFusedCell} and (5) \texttt{cuDNNLSTM}. 

To summarize, neural network researchers are confronted with a two-fold choice of framework and implementation, and identifying the fastest option helps to streamline the experimental research phase. This study aims to assist the identification process with the following goals:
\begin{itemize}
    \item Provide benchmarks for different \ac{LSTM} implementations across deep learning frameworks
    \item Use common input sizes, network configurations and cost functions as in typical \ac{ASR} scenarios
    \item Share the benchmark scripts for transparency and to help people coding up neural networks in different frameworks
\end{itemize}

\section{Related work}
The deep learning community put considerable effort into benchmarking and comparing neural-network related hardware, libraries and frameworks. A non-exhaustive list is presented followingly:
\begin{itemize}
    \item \textbf{DeepBench~\cite{deepbench}.} This project aims to benchmark basic neural network operations such as matrix multiplies and convolutions for different hardware platforms and neural network libraries such as cuDNN or MKL. In comparison, our study focuses solely on the \ac{LSTM} unit and compares on the higher abstraction level of deep learning frameworks.
    \item \textbf{DAWNBench~\cite{coleman2017dawnbench}.} This is a benchmark suite for complete end-to-end models that measures computation time and cost to train deep models to reach a certain accuracy. In contrast, our study is limited to measuring the training time per batch.
    \item \textbf{Comparative studies.} Other comparative studies such as \cite{bahrampour2016comparative} have only small sections on \acp{LSTM} and are already quite old for deep learning time scales (2016, PyTorch was not even released back then).
    \item \textbf{CNN-benchmarks.} For standard \acp{CNN} such as AlexNet~\cite{NIPS2012_4824}, VGG~\cite{Simonyan14c} or ResNet~\cite{he2016deep}, there exist several benchmark repositories \cite{soumith, u39kun}. 
    \item \textbf{TensorFlow \acp{LSTM}.} Benchmarks for the TensorFlow \ac{LSTM} variants are presented in \cite{returnn}, but deep learning frameworks other than TensorFlow are not considered.
    \item \textbf{Chainer \acp{LSTM}.} Benchmarks for the Chainer \ac{LSTM} variants are presented in \cite{chainer}, but deep learning frameworks other than Chainer are not evaluated.
\end{itemize}

\section{Setup}
The experiments cover (1) a comparison between the PyTorch, TensorFlow, Lasagne and Keras frameworks and (2) a comparison between four versions of PyTorch. The experimental details are explained in the following sections.

\subsection{Deep learning frameworks}
A summary of the frameworks, backends and CUDA/cuDNN versions is given in Table~\ref{tab:frameworks}. In total, four deep learning frameworks are involved in this comparison: (1)~PyTorch, (2)~TensorFlow, (3)~Lasagne and (4)~Keras. While PyTorch and TensorFlow can operate as standalone frameworks, the Lasagne and Keras frameworks rely on backends that handle the tensor manipulation. Keras allows for backend choice and was evaluated with the TensorFlow and Theano~\cite{theano} backends. Lasagne is limited to the Theano backend. Care was taken to use the same CUDA version 9.0 and a cuDNN 7 variant when possible. The frameworks were not compiled from source, but installed with the default \texttt{conda} or \texttt{pip} packages. Note that the development of Theano has been stopped~\cite{theano_stop}.

\begin{table*}
  \caption{Deep learning frameworks considered for evaluation. Top half: framework comparison, bottom half: PyTorch comparison}
  \label{tab:frameworks}
  \centering
  \small
  \begin{tabular}{llllll}
        \toprule
        Framework & Version & Release & Backends & CUDA & cuDNN\\
        \midrule
        PyTorch & 0.4.0 & April 2018 & - & 9.0 & 7102\\
        TensorFlow & 1.8.0 & April 2018 & - & 9.0 & 7005\\
        Lasagne & 0.2.1dev & April 2018 & Theano 1.0.1 & 9.0 & 7005\\
        Keras & 2.1.6 & April 2018 & Theano 1.0.1, TensorFlow 1.8.0 & 9.0  & 7005\\
        \midrule
        PyTorch & 0.4.0 & April 2018 & - & 9.0 & 7102\\
        PyTorch & 0.3.1post2 & February 2018 & - & 8.0 & 7005\\
        PyTorch & 0.2.0\_4 & August 2017 & - & 8.0 & 6021\\
        PyTorch & 0.1.12\_2 & May 2017 & - & 8.0 & 6021\\
  \end{tabular}
 \end{table*}

\subsection{LSTM implementations}
The benchmarks cover 10 \ac{LSTM} implementations, including the cuDNN \cite{appleyard2016optimizing} and various optimized and basic variants. The complete list of the covered \ac{LSTM} implementations is given in Table~\ref{tab:lstm_implementations}. The main differences between the implementations occur in the computation of a single time step and the realization of the loop over time, and possible optimization steps are described in \cite{appleyard2016optimizing}. While the cuDNN and optimized variants are generally faster, the basic variants are of high interest as they are easy to modify, therefore simplifying the exploration of new recurrent network cells for researchers.


    

\subsection{Network configurations}
Four network configurations were evaluated as reported in Table~\ref{tab:networks}.

\paragraph*{Input data}
The input data covers a \textit{short} sequence length (100 time steps) and a \textit{long} sequence length scenario (up to 1000 time steps). The data is randomly sampled from a normal distribution $\mathcal{N}(\mu=0, \sigma=1)$.
\begin{itemize}
    \item \textbf{Short \& fixed length.} The short input size is 64x100x123 (batch size x time steps x features) and the sequence length is fixed to 100 time steps. The target labels consist of 10 classes and 1 label is provided per sample. This setup is similar to an isolated digit recognition task on the TIDIGITS~\cite{leonard1993tidigits} data-set. \footnote{\ac{ASR}-task on TIDIGITS/isolated digit recognition, default training set (0.7 hours of speech): 123-dimensional filterbank features with 100fps, average sequence length of 98, alphabet size of 10 digits and 1 label per sample}
    
    \item \textbf{Long \& fixed length.} The long input size is 32x1000x123 and covers 1000 time steps. The target labels consist of 10 classes and 1 label is provided per sample.
    
    \item \textbf{Long \& variable length.} The sequence length is varied between 500 to 1000 time steps, resulting in an average length of 750 time steps. All 32 samples are zero-padded to the maximum length of 1000 time steps, resulting in a 32x1000x123 input size. The target labels consist of 59 output classes, and each sample is provided with a label sequence of 100 labels. The variable length setup is similar to a typical continuous speech recognition task on the \ac{WSJ}~\cite{garofalo2007csr} data-set.\footnote{\ac{ASR}-task on \ac{WSJ}/continuous speech recognition, pre-processing with EESEN \cite{miao2015eesen} on training subset si-284 (81h of speech): 123-dimensional filterbank features with 100fps, average sequence length 783, alphabet size of 59 characters and average number of characters per sample 102}. The variable length information is presented in the library specific format, i.e. as \texttt{PackedSequence} in PyTorch, as \texttt{sequence\_length} parameter of \texttt{dynamic\_rnn} in TensorFlow and as a \texttt{mask} in Lasagne. 
\end{itemize}

\paragraph*{Loss functions}
The fixed length data is classified with the cross-entropy loss function, which is integrated in all libraries. The variable length data is classified with the \ac{CTC}~\cite{graves2006connectionist} loss. For TensorFlow, the integrated \texttt{tf.nn.ctc\_loss} is used. PyTorch and Lasagne do not include \ac{CTC} loss functions, and so the respective bindings to Baidu's warp-ctc \cite{baidu_ctc} are used~\cite{pytorch-warp-ctc, theano_ctc}.

\paragraph*{Model}
All networks consist of \acp{LSTM} followed by an output projection. The \ac{LSTM} part uses either a single layer of 320 unidirectional \ac{LSTM} units, or four layers of bidirectional \acp{LSTM} with 320 units per direction. When using the cross entropy loss, the output layer consists of 10 dense units that operate on the final time step output of the last \ac{LSTM} layer. In contrast, the output layer of the CTC loss variant uses 59 dense units that operate on the complete sequence output of the last \ac{LSTM} layer.

 \begin{table}[h]
\begin{adjustbox}{center}
\begin{threeparttable}
  \caption{LSTM implementations considered for evaluation. Click on the name for a hyperlink to the documentation.}
  \label{tab:lstm_implementations}
    \def\arraystretch{1.5}
    \newcolumntype{C}{>{\centering\arraybackslash}m{0.5cm}}
    \newcolumntype{D}{>{\centering\arraybackslash}m{.0cm}}   
    \footnotesize

  \begin{tabularx}{1.2\textwidth}{m{1.5cm}m{3.0cm}CCCCm{7.5cm}}
        \toprule
        Framework & Name & \rot{1x320/CE-short} & \rot{1x320/CE-long} & \rot{4x320/CE-long} & \rot{4x320/CTC-long} & Detail\\
        \midrule
        PyTorch & \texttt{\href{https://github.com/stefbraun/rnn_benchmarks/blob/master/1x320-LSTM/bench_pytorch_LSTMCell-basic.py}{LSTMCell-basic}} & \cmark &\cmark & \xmark\tnote{1} & \xmark\tnote{1} & Custom code, pure PyTorch implementation, easy to modify. Loop over time with Python \texttt{for} loop\\
        PyTorch & \texttt{\href{http://pytorch.org/docs/stable/nn.html?highlight=lstmcell\#torch.nn.LSTMCell}{LSTMCell-fused}}\tnote{2} & \cmark & \cmark & \xmark\tnote{1} & \xmark\tnote{1} &  LSTM with optimized kernel for single time steps. Loop over time with Python \texttt{for} loop\\
        PyTorch & \texttt{\href{http://pytorch.org/docs/stable/nn.html?highlight=lstm\#torch.nn.LSTM}{cuDNNLSTM}}\tnote{3} & \cmark & \cmark & \cmark & \cmark & Wrapper to cuDNN LSTM implementation \cite{appleyard2016optimizing}\\
        TensorFlow & \texttt{\href{https://www.tensorflow.org/versions/r1.8/api_docs/python/tf/contrib/rnn/LSTMCell}{LSTMCell}} & \cmark & \cmark & \cmark & \cmark & Pure TensorFlow implementation, easy to modify. Loop over time with \texttt{tf.while\_loop}. Uses \texttt{dynamic\_rnn}\\ 
        TensorFlow & \texttt{\href{https://www.tensorflow.org/versions/r1.8/api_docs/python/tf/contrib/rnn/LSTMBlockCell}{LSTMBlockCell}} & \cmark & \cmark & \cmark & \cmark & Optimized LSTM with single operation per time-step. Loop over time with \texttt{tf.while\_loop}. Uses \texttt{dynamic\_rnn}\\
        TensorFlow & \texttt{\href{https://www.tensorflow.org/versions/r1.8/api_docs/python/tf/contrib/rnn/LSTMBlockFusedCell}{LSTMBlockFusedCell}} & \cmark & \cmark & \xmark\tnote{1} & \xmark\tnote{1} & Optimized LSTM with single operation over all time steps. Loop over time is part of the operation.\\
        TensorFlow & \texttt{\href{https://www.tensorflow.org/versions/r1.8/api_docs/python/tf/contrib/cudnn_rnn/CudnnLSTM}{cuDNNLSTM}} & \cmark & \cmark & \cmark & \xmark\tnote{4} & Wrapper to cuDNN LSTM implementation \cite{appleyard2016optimizing}\\
        Lasagne & \texttt{\href{http://lasagne.readthedocs.io/en/latest/modules/layers/recurrent.html?highlight=gru\#lasagne.layers.LSTMLayer}{LSTMLayer}} & \cmark & \cmark & \cmark & \cmark & Pure Theano implementation, easy to modify. Loop over time with \texttt{theano.scan}\\
        Keras & \texttt{\href{https://keras.io/layers/recurrent/\#lstm}{LSTM}} & \cmark & \cmark & \xmark\tnote{1} & \xmark\tnote{1} & Pure Theano/TensorFlow implementation, easy to modify. Loop over time with \texttt{theano.scan} or \texttt{tf.while\_loop}\\
        Keras & \texttt{\href{https://keras.io/layers/recurrent/\#cudnnlstm}{cuDNNLSTM}} & \cmark &\cmark & \xmark\tnote{1} & \xmark\tnote{1} & Wrapper to cuDNN LSTM implementation~\cite{appleyard2016optimizing}\tnote{5}\\
        \bottomrule
  \end{tabularx}
  \begin{tablenotes}
  \item [1] no helper function to create multi-layer networks
  \item [2] renamed from original name \texttt{LSTMCell} for easier disambiguation
  \item [3] renamed from original name \texttt{LSTM} for easier disambiguation
  \item [4] no support for variable sequence lengths
  \item [5] only available with TensorFlow backend
  
  \end{tablenotes}
 \end{threeparttable}
\end{adjustbox} 

\bigskip
  \caption{Neural network configurations considered for evaluation}
  \label{tab:networks}
  \centering
  \begin{adjustbox}{center}

  \resizebox{1.2\textwidth}{!}{
  \begin{tabular}{lcccccc}
        \toprule
        Name & Layers x LSTM units & Output & Loss & Input [NxTxC] & Sequence length & Labels per sample\\
        \midrule
        1x320/CE-short & 1x320 unidirectional & 10 Dense & cross entropy & 64x100x123 & fixed $\in\{100\}$ & 1\\
        1x320/CE-long & 1x320 unidirectional & 10 Dense & cross entropy & 32x1000x123 & fixed $\in\{1000\}$ & 1\\
        4x320/CE-long & 4x320 bidirectional & 10 Dense & cross entropy & 32x1000x123 & fixed $\in\{1000\}$ & 1 \\
        4x320/CTC-long & 4x320 bidirectional & 59 Dense & CTC & 32x1000x123 & variable $\in \{500, ..., 1000\}$ & 100\\
        \midrule
        
  \end{tabular}
  }
  \end{adjustbox}
\end{table}

\paragraph*{Optimizer} All benchmarks use the framework-specific default implementation of the widely used ADAM optimizer~\cite{kingma2014adam}.

\subsection{Measurements}
The reported timings reflect the mean and standard deviation of the time needed to fully process one batch, including the forward and backward pass. The benchmarks were carried out on a machine with a Xeon W-2195 CPU (Skylake architecture\footnote{This CPU scales its clock frequency between 2.3GHz to 4.3Ghz. In order to reduce the fluctuation of the CPU frequency, the number of available CPU cores was restricted to 4 such that all 4 cores maintained 4.0GHz to 4.3GHz clock frequency}), a NVIDIA GTX 1080 Founders Edition graphics card (fan speed @ 100\% to avoid thermal throttling) and Ubuntu 16.04 operating system (CPU frequency governor in \texttt{performance} mode\footnote{By default, Ubuntu 16.04 uses the mode \texttt{powersave} which significantly decreased performance during the benchmarks}). The measurements were conducted over 500 iterations and the first 100 iterations were considered as warm-up and therefore discarded. 

\subsection{Known limitations}
The limitations reflect the current state of the benchmark code, and may be fixed in future versions. Further issues can be reported in the github repository.
\begin{itemize}
    \item The benchmark scripts are carefully written, but not optimized to squeeze that last bit of performance out of them. They should reflect typical day-to-day research applications.
    \item Due to time constraints, only the 1x320 LSTM benchmark covers all considered frameworks. For the multi-layer 4x320 networks, only implementations that provided helper functions to create stacked bidirectional networks were evaluated. An exemption of this rule was made for Lasagne, in order to include a Theano-based contender for this scenario.
    \item The TensorFlow benchmarks use the \texttt{feed\_dict} input method that is simple to implement, but slower than the \texttt{tf.data} API~\cite{feed_dict}. Implementing a high performance input pipeline in TensorFlow is not trivial, and only the \texttt{feed\_dict} approach allowed for a similar implementation complexity as in the PyTorch and Lasagne cases.
    \item The TensorFlow \texttt{cuDNNLSTM} was not tested with variable length data as it does not support such input \cite{6633}.
    \item The TensorFlow benchmark uses the integrated \texttt{tf.nn.ctc\_loss} instead of the warp-ctc library, even though there is a TensorFlow binding available \cite{baidu_ctc}. The performance difference has not been measured.
    \item PyTorch 0.4.0 merged the \texttt{Tensor} and \texttt{Variable} classes and does not need the \texttt{Variable} wrapper anymore. The \texttt{Variable} wrapper has a negligible performance impact on version 0.4.0, but is required for older PyTorch releases in the PyTorch version comparison.
\end{itemize}

\section{Results}
The complete set of results is given in Table~\ref{tab:results} and the most important findings are summarized below.

\begin{itemize}
    \item \textbf{Fastest \ac{LSTM} implementation.} The \texttt{cuDNNLSTM} is the overall fastest LSTM implementation, for any input size and network configuration. It is up to 7.2x faster than the slowest implementation (Keras/TensorFlow \texttt{LSTM
    }, 1x320/CE-long). PyTorch, TensorFlow and Keras provide wrappers to the cuDNN LSTM implementation and the speed difference between frameworks is small (after all, they are wrapping the same implementation).
    
    \item \textbf{Optimized LSTM implementations.} When considering only optimized \ac{LSTM} implementations other than \texttt{cuDNNLSTM}, then the TensorFlow \texttt{LSTMBlockFusedCell} is the fastest variant: it is 1.3x faster than PyTorch \texttt{LSTMCell-fused} and 3.4x faster than TensorFlow \texttt{LSTMBlockCell} (1x320/CE-long). 
    The fused variants are slower than the \texttt{cuDNNLSTM} but faster than the more flexible alternatives. 
    
    \item \textbf{Basic/Flexible LSTM implementations.} When considering only flexible implementations that are easy to modify, then Lasagne \texttt{LSTMLayer}, Keras/Theano \texttt{LSTM} and PyTorch \texttt{LSTMCell-basic} are the fastest variants with negligible speed difference. All three train 1.6x faster than TensorFlow \texttt{LSTMCell}, and 1.8x faster than Keras/TensorFlow \texttt{LSTM} (1x320/CE-long). For Keras, the Theano backend is faster than the TensorFlow backend.

    \item \textbf{100 vs 1000 time steps.} The results for the short and and long input sequences allow for similar conclusions, with the main exception being that the spread of training time increases and the fastest implementation is 7.2x (1x320/CE-long) vs. 5.1x (1x320/CE-short) faster than the slowest.
    
    \item \textbf{Quad-layer networks.} The \texttt{cuDNNLSTM} wrappers provided by TensorFlow and PyTorch are the fastest implementation and deliver the same training speed, and they are between 4.7x to 7.0x faster than the networks built with Lasagne \texttt{LSTMLayer} and TensorFlow \texttt{LSTMBlockCell} / \texttt{LSTMCell} (4x320/CE-long and 4x320/CTC-long). 
    
    \item \textbf{Fixed vs. variable sequence length.} Going from 4x320/CE-long to 4x320/CTC-long (fixed vs. variable sequence length, cross entropy vs. CTC loss function) slows down training by a factor of 1.1x (PyTorch \texttt{cuDNNLSTM}) to 1.2x (Lasagne \texttt{LSTMLayer}). The TensorFlow implementations stabilize at roughly the same level. 
    
    \item \textbf{PyTorch versions.} The two most recent versions 0.4.0 and 0.3.1post2 are significantly faster than the older versions, especially for the flexible \texttt{LSTMCell-basic} where the speedup is up to 2.2x (1x320/CE-long). The 4x320/CTC-long benchmark is very slow for version 0.2.0\_4 which suffered from a slow implementation for sequences of variable length that is fixed in newer versions \cite{packed_sequence}. The older PyTorch versions tend to produce larger standard deviations in processing time, which is especially visible on the 1x320/CE-short test.
    
\end{itemize}

\section{Conclusion}
This study evaluated the training time of various \ac{LSTM} implementations in the PyTorch, TensorFlow, Lasagne and Keras deep learning frameworks. The following conlusions are drawn:
\begin{itemize}
    \item The presented \ac{LSTM} benchmarks show that an informed choice of deep learning framework and \ac{LSTM} implementation may increase training speed by up to 7.2x on standard input sizes from \ac{ASR}. 
    \item The overall fastest \ac{LSTM} implementation is the well-optimized cuDNN variant provided from NVIDIA which is easily accessible in PyTorch, TensorFlow and Keras, and the training speed is similar across frameworks. 
    \item When comparing deep learning frameworks and less optimized, but more customizable \ac{LSTM} implementations, then PyTorch trains 1.5x to 1.6x faster than TensorFlow and Lasagne trains between 1.3x to 1.6x faster than TensorFlow. Keras is between 1.5x to 1.7 faster than TensorFlow when using the Theano backend, but 1.1x slower than TensorFlow when using the TensorFlow backend.
    \item The comparison of PyTorch versions showed that for PyTorch users it is a good idea to update to the most recent version.    
\end{itemize}
On a final note, the interested reader is invited to visit the github page and run his own benchmarks with custom network and input sizes.

\subsubsection*{Acknowledgments}
I thank my PhD supervisor Shih-Chii Liu for support and encouragement in the benchmarking endeavours. I also thank the Sensors group from the Institute of Neuroinformatics, University of Zurich / ETH Zurich for feedback and discussions. This work was partially supported by Samsung Advanced Institute of Technology.

\begin{table*}
\footnotesize
\centering
\begin{adjustbox}{center, max width=0.85\textwidth}
\tabcolsep=0cm
\begin{threeparttable}
    
  \caption{Benchmarking results: mean and standard deviation of training time per batch in [ms]}
  \label{tab:results}
  \centering

  \begin{tabular}{cc}

  \toprule
        Framework comparison & PyTorch comparison\\
        \midrule
        
        \includegraphics[trim={0cm 0cm 0cm 0.5cm}, clip=true, width=0.75\linewidth]{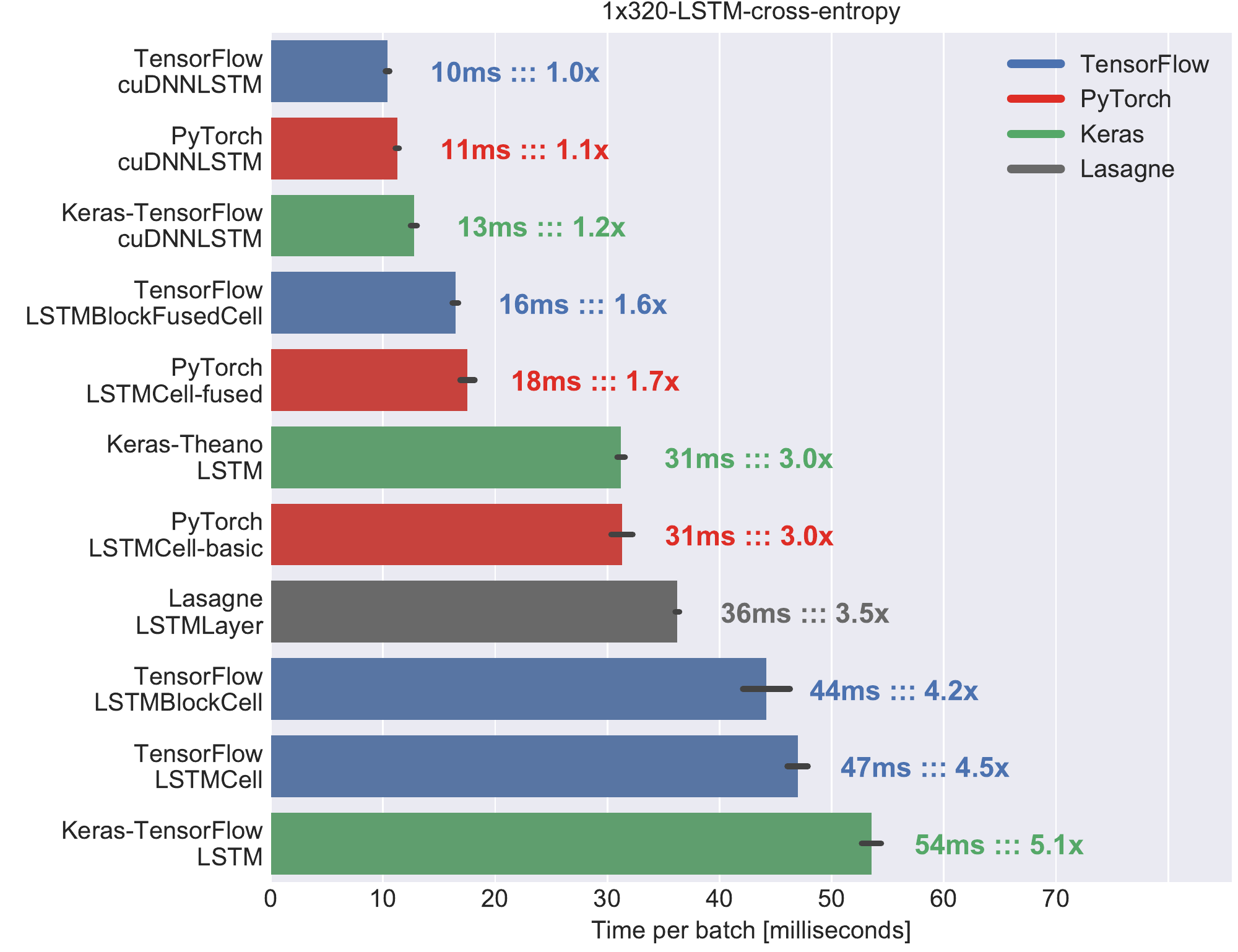} & \includegraphics[trim={0cm 0cm 0cm 0.5cm}, clip=true, width=0.75\linewidth]{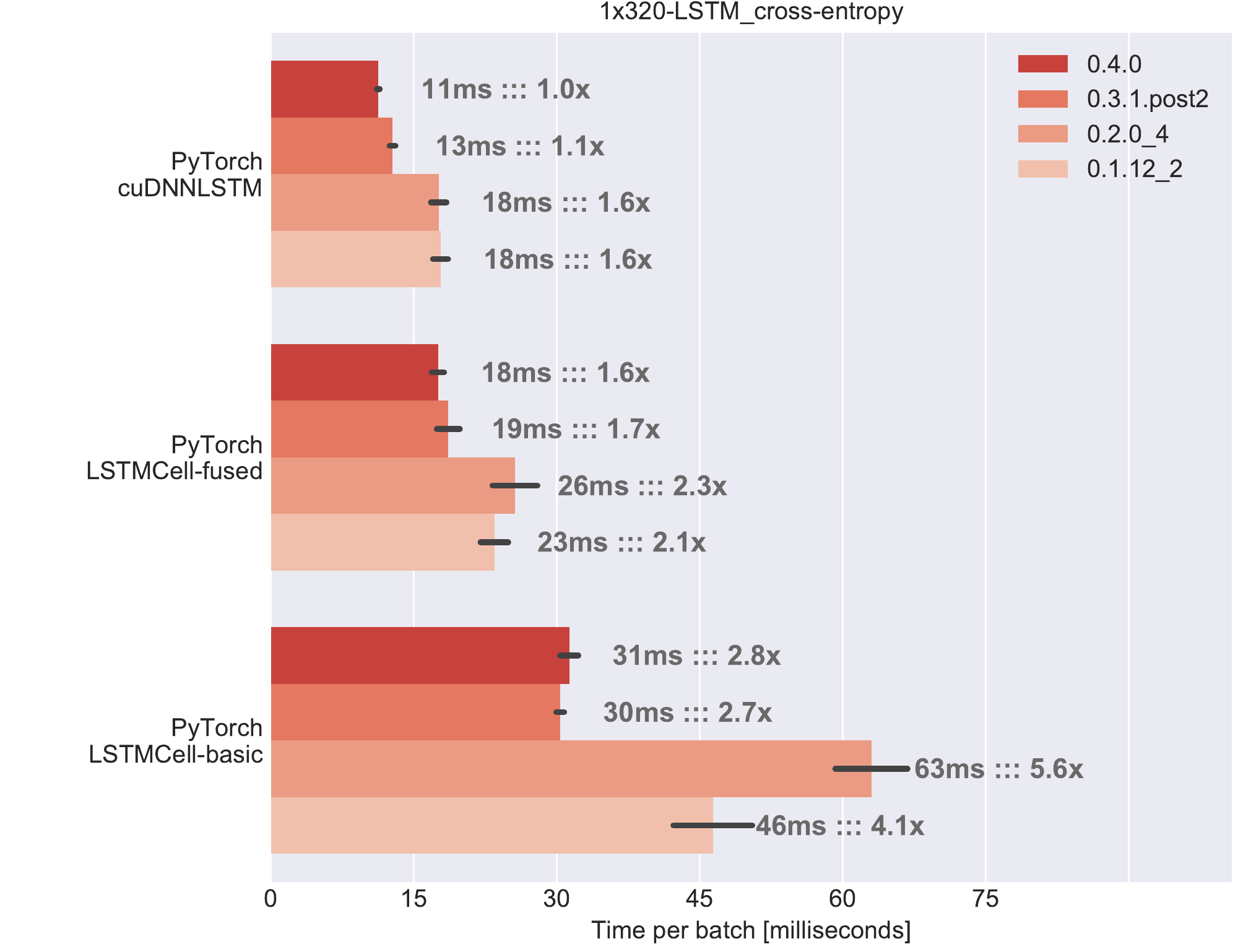} \\
        \multicolumn{2}{c}{(a) \textbf{1x320/CE-short} ::: 1x320 unidirectional LSTM ::: cross entropy loss ::: fixed sequence length ::: input 64x100x123 (NxTxC)\tnote{1}}\\
        \midrule
        
        \includegraphics[trim={0cm 0cm 0cm 0.5cm}, clip=true, width=0.75\linewidth]{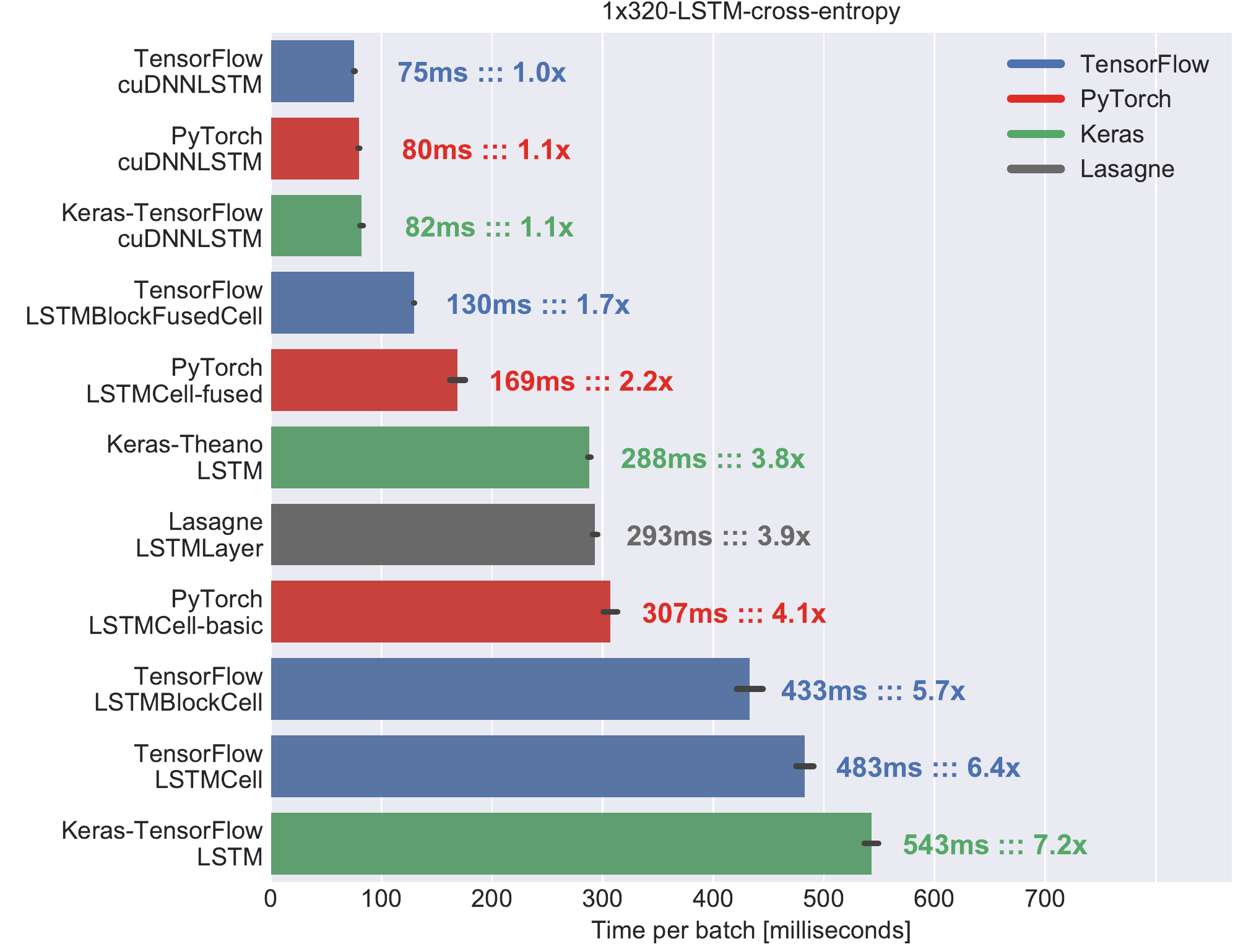} & \includegraphics[trim={0cm 0cm 0cm 0.5cm}, clip=true, width=0.75\linewidth]{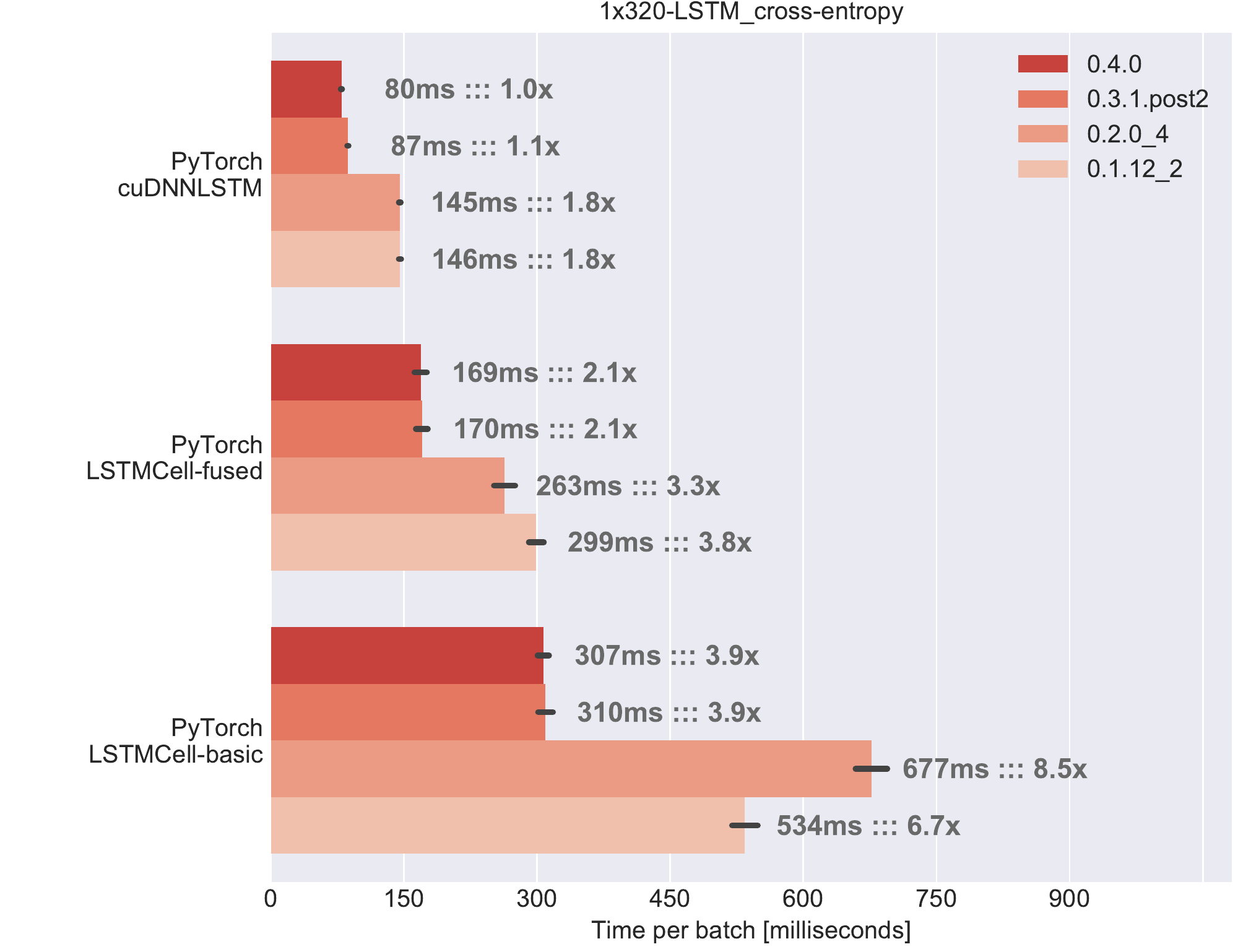} \\
        \multicolumn{2}{c}{(b) \textbf{1x320/CE-long} ::: 1x320 unidirectional LSTM ::: cross entropy loss ::: fixed sequence length ::: input 32x1000x123 (NxTxC)\tnote{1}}\\
        \midrule

        \includegraphics[trim={0cm 0cm 0cm 0.5cm}, clip=true, width=0.75\linewidth]{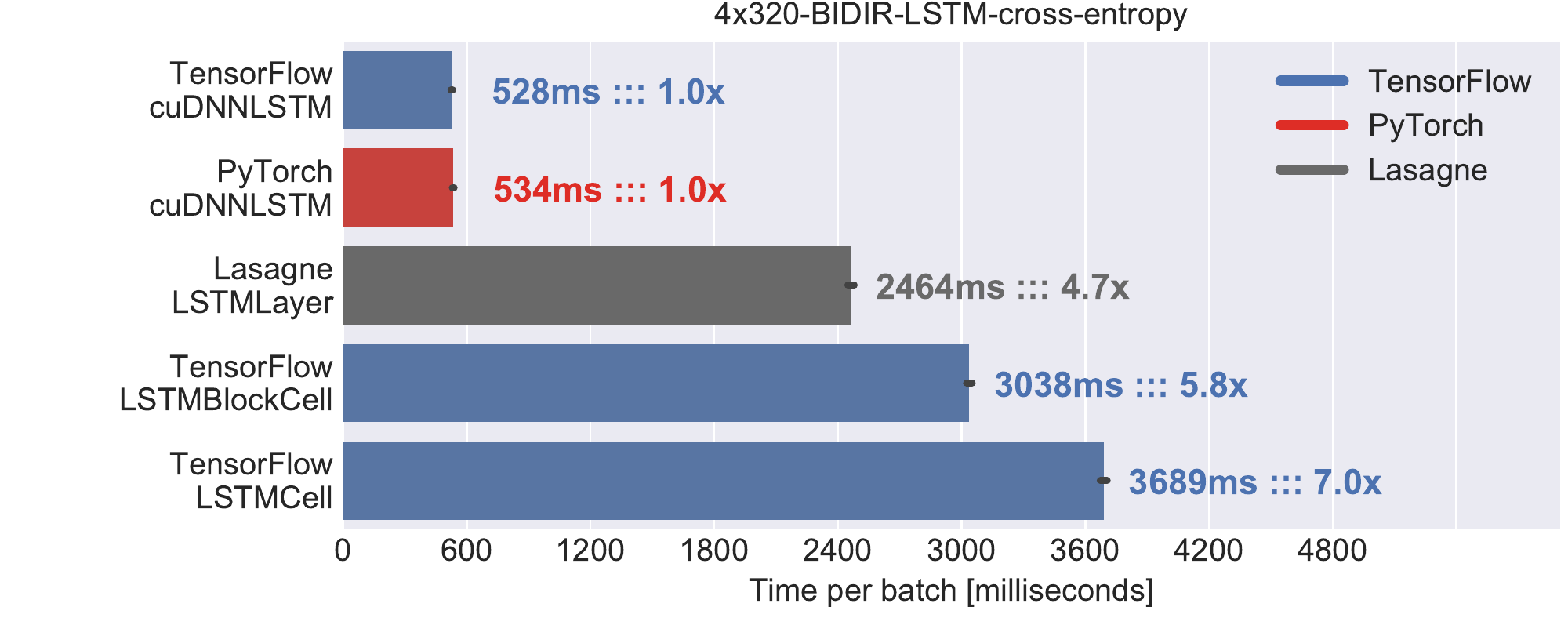} & \includegraphics[trim={0cm 0cm 0cm 0.5cm}, clip=true, width=0.75\linewidth]{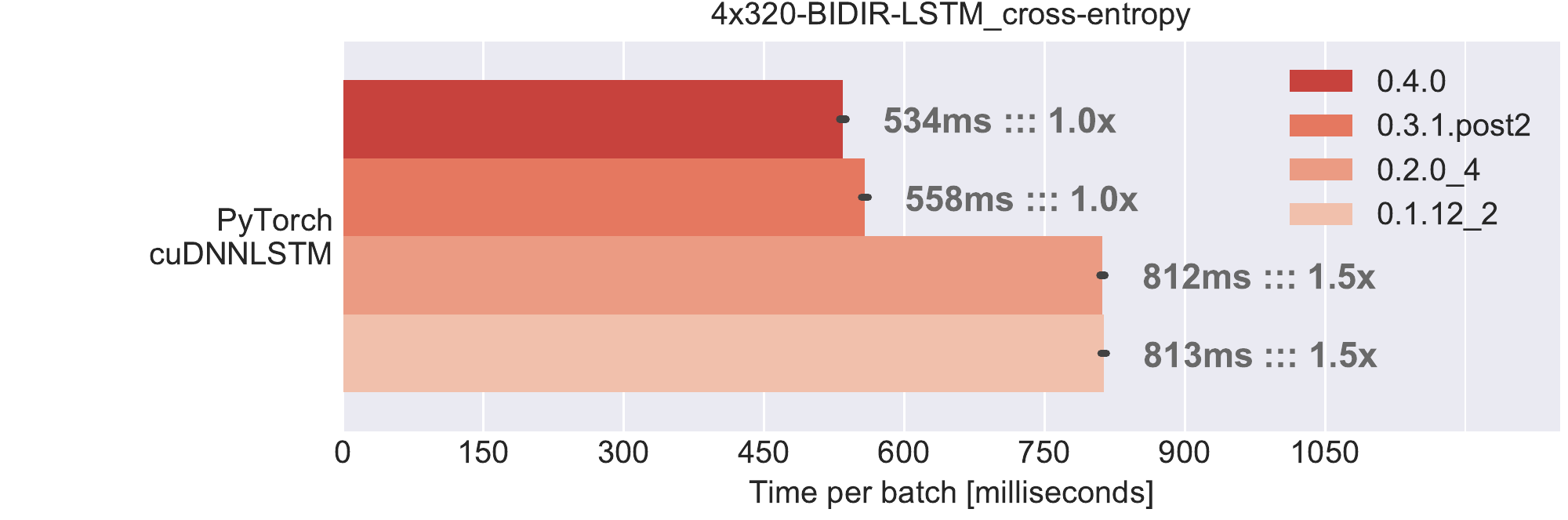} \\
        \multicolumn{2}{c}{(c) \textbf{4x320/CE-long} ::: 4x320 bidirectional LSTM ::: cross entropy loss ::: fixed sequence length ::: input 32x1000x123 (NxTxC)\tnote{1}}\\
        \midrule
        
        \includegraphics[trim={0cm 0cm 0cm 0.5cm}, clip=true, width=0.75\linewidth]{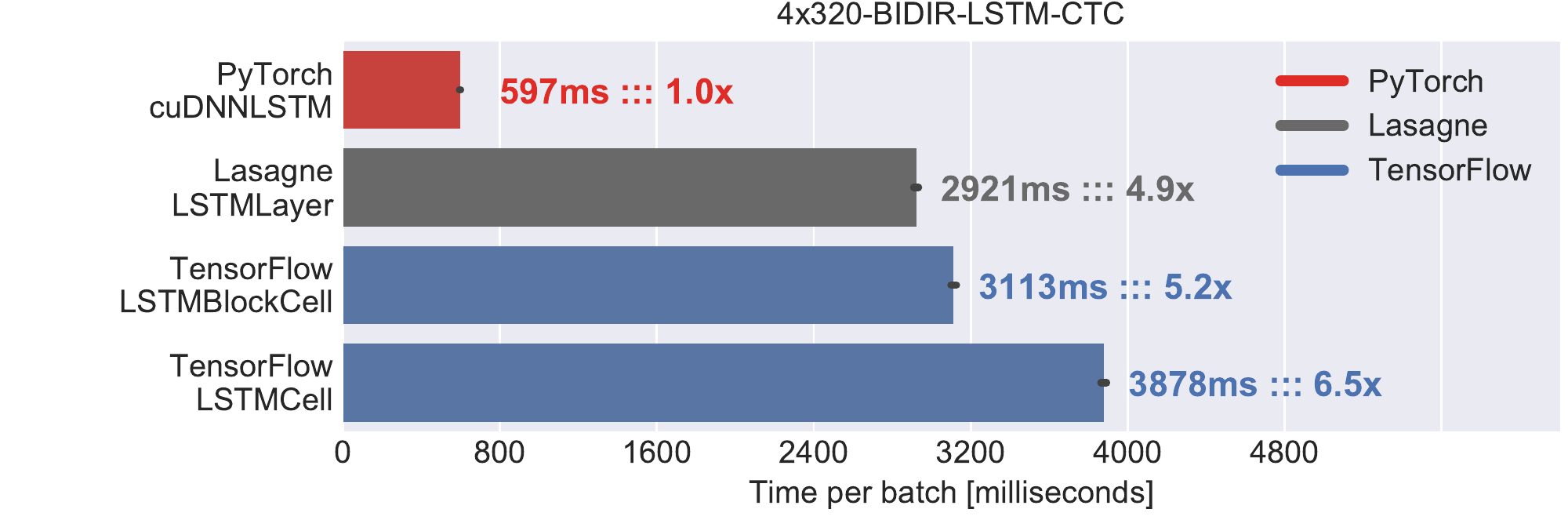} & \includegraphics[trim={0cm 0cm 0cm 0.5cm}, clip=true, width=0.75\linewidth]{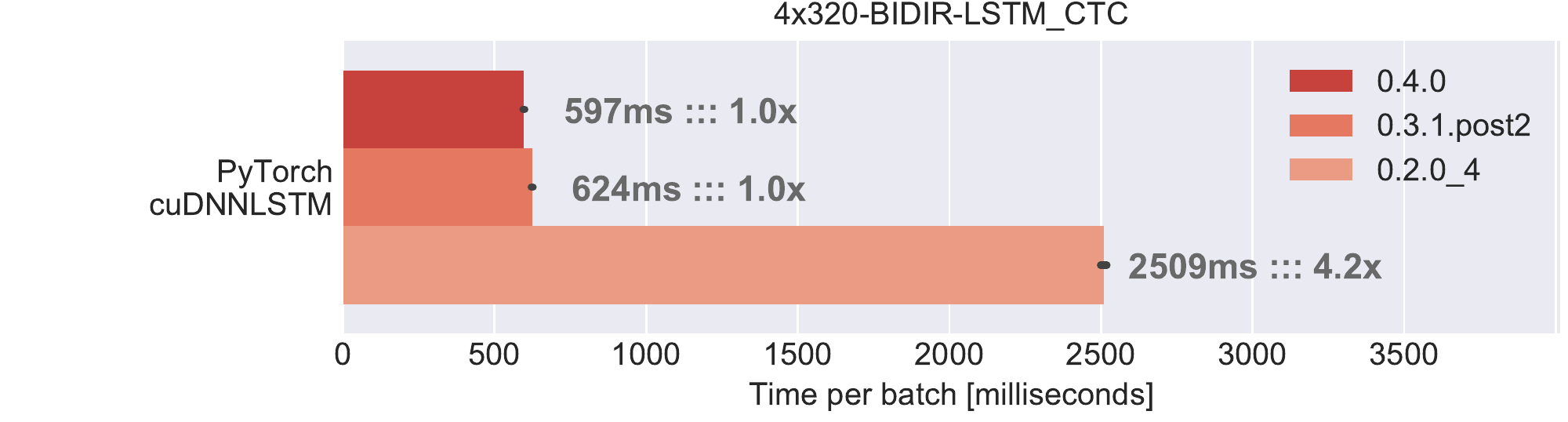} \\
        \multicolumn{2}{c}{(d) \textbf{4x320/CTC-long} ::: 4x320 bidirectional LSTM ::: CTC loss ::: variable sequence length ::: input 32x1000x123 (NxTxC)\tnote{1,2}}\\
    \bottomrule
  \end{tabular}
  \begin{tablenotes}
  \item [1] N = batch size, T = time steps, C =feature channels
  \item [2] CTC experiment for pytorch 0.1.12\_2 omitted due to warp-ctc compilation issues
  \end{tablenotes}
\end{threeparttable}
\end{adjustbox}
\end{table*}

\clearpage
\small
\bibliographystyle{ieeetr}

\bibliography{main}







\end{document}